\documentclass{article}
\usepackage{icmc2025_paper_template}
\usepackage{times}
\usepackage{ifpdf}
\usepackage{soul}
\usepackage[english]{babel}

\usepackage{amsmath} % popular packages from Am. Math. Soc. Please use the 
\usepackage{algorithm}
\usepackage[noend]{algpseudocode}
\def\papertitle{A Graph Engine for Guitar Chord-Tone Soloing Education}
\def\firstauthor{Matthew Keating}
\def\secondauthor{Michael Casey}
\newif\ifpdf
\ifx\pdfoutput\relax
\else
   \ifcase\pdfoutput
      \pdffalse
   \else
      \pdftrue
  \fi
\fi

\ifpdf % compiling with pdflatex
  \usepackage[pdftex,
    pdftitle={\papertitle},
    pdfauthor={\firstauthor, \secondauthor},
    bookmarksnumbered, % use section numbers with bookmarks
    pdfstartview=XYZ % start with zoom=100% instead of full screen; 
   ]{hyperref}
  \usepackage[pdftex]{graphicx}
  \graphicspath{{./figures/}}
  \DeclareGraphicsExtensions{.pdf,.jpeg,.png}

  \usepackage[figure,table]{hypcap}

\else % compiling with latex
  \usepackage[dvips,
    bookmarksnumbered, % use section numbers with bookmarks
    pdfstartview=XYZ % start with zoom=100% instead of full screen
  ]{hyperref}  % hyperrefs are active in the pdf file after conversion

  \usepackage[dvips]{epsfig,graphicx}
  \graphicspath{{./figures/}}
  \DeclareGraphicsExtensions{.eps}

  \usepackage[figure,table]{hypcap}
\fi

\hypersetup{
    colorlinks,%
    citecolor=black,%
    filecolor=black,%
    linkcolor=black,%
    urlcolor=black
}

\title{\papertitle}

\twoauthors
   {0.5in}
   {\firstauthor} {Dartmouth College \\  %
    {\tt \href{mailto:matthew.b.keating.23@dartmouth.edu}{matthew.b.keating.23@dartmouth.edu}}}
   {\secondauthor} {Dartmouth College \\  %
     {\tt \href{mailto:michael.a.casey@dartmouth.edu}{michael.a.casey@dartmouth.edu}}}

\begin{document}
\capstartfalse
\maketitle
\capstarttrue
\begin{abstract}
We present a graph-based engine for computing chord tone soloing suggestions for guitar students. Chord tone soloing is a fundamental practice for improvising over a chord progression, where the instrumentalist uses only the notes contained in the current chord. This practice is a building block for all advanced jazz guitar theory but is difficult to learn and practice. First, we discuss methods for generating chord-tone arpeggios. Next, we construct a weighted graph where each node represents a chord-tone arpeggio for a chord in the progression. Then, we calculate the edge weight between each consecutive chord's nodes in terms of optimal transition tones. We then find the shortest path through this graph and reconstruct a chord-tone soloing line. Finally, we discuss a user-friendly system to handle input and output to this engine for guitar students to practice chord tone soloing.
\end{abstract}

\section{Introduction}\label{sec:introduction}
Improvised soloing is a core part of playing jazz music. Traditional jazz bebop standards have a set chord progression the band plays while the soloist improvises musical lines over the chords. Part of learning to improvise on a jazz chord progression is to use chord-tone arpeggios that fit each chord, a process known as chord-tone soloing. A chord tone is defined as a note in a chord. For example, the chord Amin7 comprises the notes A, C, E, and G, so these four notes are all chord tones for Amin7. Learning chord tone arpeggios is particularly difficult on guitar because the guitar is a transposition instrument and has multiple correct ways of playing any one note.

The standard guitar has six strings and approximately 22 frets. Without using extended techniques such as bends, each fretted note on the guitar plays a note on the 12-tone chromatic scale. Therefore, the guitar has about 132 possible fretted positions and 11 different positions for each note in the 12-tone scale. A chord tone arpeggio for a four-note chord, standard for jazz chord voicings, is four notes and one-third of the 12-tone chromatic scale. So, approximately 44 positions on the guitar neck play a chord tone for any four-note chord.

Chord tone soloing becomes more complicated when expanded to a whole chord progression. The soloist must transition to playing the next chord's tones when the current chord changes. Generally, the best practice for moving between chord-tone arpeggios minimizes the distance traveled between the transition notes. Therefore, a guitarist must know the chord tone arpeggios in all positions on the guitar fretboard to select the chord tone positions with the best transition notes.

Learning these chord tone techniques on jazz guitar is a huge task. Some online resources exist such as textbooks and instructional videos, but many of these resources are behind paywalls. Furthermore, these resources are static content for a specific learning case, such as soloing over a well-known chord progression, and do not extend to every chord progression a student may encounter. In this paper, we detail a system that can generate chord-tone soloing examples for any chord progression.

Generating educational material for guitar instruction is a growing topic in computer music. Keating et al. used long short-term memory models to generate guitar chord voicing examples for jazz chord progressions \cite{Keating:24}. We expand on this model by offering an approach for chord tone solos and constraining the model to always produce a meaningful result by using simple graph traversals. Our weighted graph-based engine is similar to a Hidden Markov Model (HMM) approach, common for tablature generation models such as Hori et al. which generates tablature from staff notation \cite{Hori:13} and in piano fingering estimation models such as Nakamura et al. \cite{Nakamura:2020}. HMMs use similar state representations, which we use to represent different chord tone arpeggios, but have probabilities to represent state change and are data-driven. In contrast, our model has linear weights and is weighted with certain quantifiable metrics.

Other related research includes improvisation generation models for creating music for performance. Frieler et al. created a data-driven improvisation generation model and tested performance with human listeners \cite{Frieler:22}. Pachet's Continuator uses a Markov model to generate plausible notes for improvisation \cite{Pachet:03}. These models have similar goals to our model insofar as they generate improvised solos, but they are focused on machine-generated music for performance rather than music education, and their output is not mapped onto a guitar. Further related work includes writing figured bass lines with rule-based algorithms \cite{Wead:2007, Yaolong:2020}. Our work employs a similar strategy to codify known musical rules for line generation.

\section{Graph Engine}\label{sec:engine}

Our engine for generating guitar chord tone suggestions relies on a weighted graph construction and traversal.\footnote{Engine code: \href{https://github.com/mbkeating/chord-tone-generator}{https://github.com/mbkeating/chord-tone-generator}} Each node in this graph represents a set of positions on the guitar that play a chord's tones. Each edge describes the movement between chords in a progression. We query this graph by traversing the shortest path and assembling the sets of notes in an optimized order. The resulting list of notes is a single-note line to play on guitar which follows the given chord progression.

\subsection{Generating guitar chord tones}

Guitar chord tones are generated by starting at the chord's root note in some fretboard position and playing each interval in the chord type. For example, an Amin7 chord has the tones A, C, E, and G. A guitarist can start on the fifth fret of the low E string and play the A, then play the C on the eighth fret of the low E string, the E on the seventh fret of the A string, and the G on the fifth fret of the D string. Alternatively, the guitarist can start the arpeggio on a different root note position, giving an equally valid set of chord tones.

We could use a set of hard-coded arpeggios to give the model an explicit set of chord-tone arpeggios. The standard arpeggios for most chord shapes are documented in jazz guitar textbooks and online articles. These chord-tone arpeggios can be recorded, indexed by their chord name, and queried when constructing a graph.

We can also use a rules-based algorithm to generate these arpeggios. For example, the one used in our preliminary study took a preferred fret stretch constant $D$, where the user does not want to stretch their fingers more than $D$ frets to reach the arpeggio. We precomputed a 2D array of all notes on a six-string 22-fret guitar $G[][]$ where $G[0][0]=E$ for the low E. Then, given each chord root note and type (where a type implies the intervals/thirds stacking ex. min7 = [3, 4, 3] or minor 3, major 3, minor 3) we iterated over all root note positions and recursively found arpeggios by moving right on the fretboard (no more than $D$ away from the root position) or up the strings. This approach lacks the expert guitarist knowledge that hard-coded arpeggios encode. However, we chose this approach for our preliminary study because the generation works well enough to demonstrate the engine and does not require an expert-curated dataset.

\subsection{Constructing the chord tone graph}

Our graph is formed with two types of objects: nodes and edges. Each node has a value tracking chord tones or NULL for a special node (source or sink). Each edge has a start node, an end node, and a numerical weight tracking the score between the two connected nodes.

A graph construction starts with receiving the desired chord progression as a framework and the desired notes per measure (NPM). For example, a user could input the chord progression Amin7, D7 with 4 notes per measure. First, we must create a source node and a sink node. This graph will be traversed to find the shortest path from source to sink. Next, we iterate through each chord in the progression and generate all possible NPM-length chord tone patterns. We construct a graph node for each of these NPM-length chord tone patterns. Then we create an edge between each node in a chord and every node in the next chord. We create an edge from the source node to every chord tone node representing the first chord, and an edge from every chord tone node representing the last node to the sink node.

The final part of the graph construction is the edge weight calculation. First, we give any edge that connects to the source or sink nodes a weight of 0. Then, we calculate the edge weights between any two chord tone nodes. An edge weight must be a function of both nodes. This edge weight should be minimized when the transition between two adjacent chord tone sets is the best by musical definition. Traditionally in jazz guitar, the best musical transition between two sets of chord tones is one in which there is some tone in both sets that is equal or very close, usually a semitone or full tone away.\footnote{In fact, we can prove that for any two four-tone sequences on a 12-tone scale with guitar, the greatest transition distance between the two sequences is a minor third away. Quick proof sketch: the greatest distance is created between a chromatic four-tone scale for both chords. The first chromatic scale will take four tones of the 12 possible tones, giving eight remaining tones. Therefore, because the 12-tone scale is modular on guitar, the remaining four tones will have at most two tones between the first chord - a minor third. } So, we use a simple edge weight calculation that gives the minimum transition note distance between two chord tone sets. For chord tone sets $C_1$ and $C_2$, this weight calculation is
\begin{equation}
w(C_1, C_2) = \min_{i \in C_1, j \in C_2} |i - j|
\end{equation}
Where the minus operator represents the fret-wise distance on a guitar between two notes. If there are no notes on the same fret, we can give a penalty for switching strings in the fret-wise minus operator. Our engine's penalty counts a string change as twice a fret change, but this constant is largely left to the guitarist's tendencies.\footnote{If the constant is greater, the engine will generate fewer examples that skip strings when finding transition tones between arpeggios.}

\begin{figure}[h]
\centering
\includegraphics[width=0.9\columnwidth]{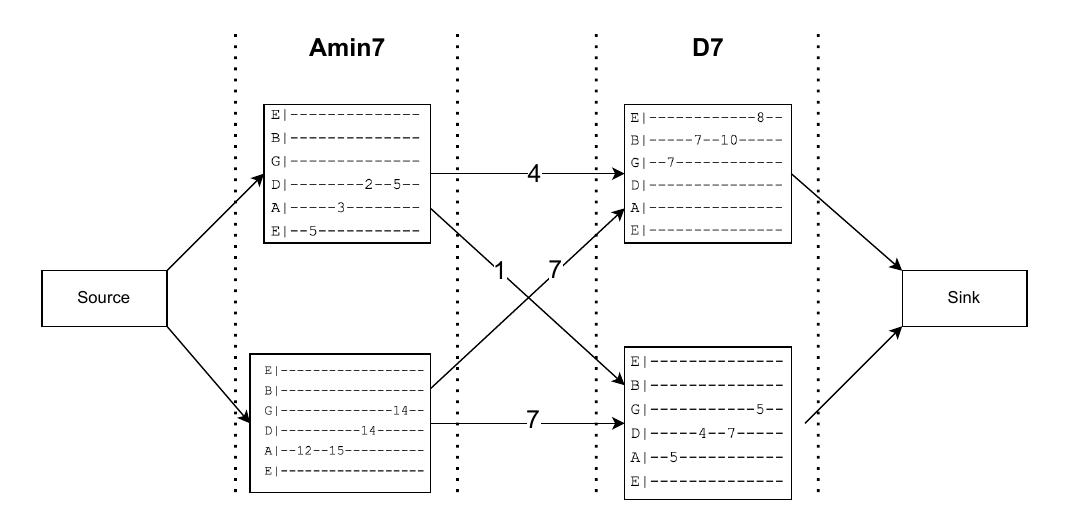}
\caption{A constructed graph with the chord progression Amin7 to D7. Each chord has two arpeggio options rendered as tablature. Each edge weight is calculated based on the simple edge weight calculation in equation (1).}
\end{figure}

Figure 1 shows an example graph for this construction for the chord progression Amin7 to D7 (traditional jazz 2-5 in G major). The possible chord tone arpeggio nodes are shortened to two nodes per chord to compact the figure, and there are more chord tone arpeggio possibilities in a programmatic construction.

The main strength of this graph model is the extensibility of the weight. The weighting on the strength of transition tone used in this graph construction is relatively simple and may not capture all the intricacies of chord tone soloing. We can also create a weight function that factors in other metrics. For example, we can calculate how far the guitarist must move their hands to form each complete arpeggio shape, which could increase the playing speed. Or, in a user-facing system, a user may "like" or "dislike" any chord arpeggio based on personal preference or familiarity, which could also be quantified and factored into the weight. These metrics can be combined linearly to allow for a complex weight. For instance, for two quantifiable metrics $w_1(C_1, C_2)$ and $w_2(C_1, C_2)$ and corresponding metric importance constants (calculated by the model engineer) $a$ and $b$, we can have a weight calculation with three metrics of varying importance by finding
\begin{equation}
w(C_1, C_2) = a w_1(C_1, C_2) + b w_2(C_1, C_2)
\end{equation}

Where a higher importance constant makes a metric affect the weight more. For our current purposes, Equation (1) is a sufficient weight calculation.

\subsection{Querying the chord tone graph}

We query the constructed graph by performing a shortest path search of the graph starting at the source node and ending at the sink node. We use Dijkstra's algorithm to compute the shortest path because our graph is weighted \cite{Dijkstra:59}. Dijkstra's algorithm guarantees that the path has the lowest edge weights, globally minimizing our chord tone arpeggio distance constraints. We reconstruct the path from source to sink as a list of chord tone values.

Each chord tone object is unordered, so we employ the following procedure to order the tones. Create an ordered list to store all the arranged chord tones. Iterate through the list of chord tones. For each pair of chord tones, find the transition tones (with the minimal fret-wise distance). Assign those transition tones to consecutive slots in the arranged chord tones. Fill in the gaps by randomly arranging the remaining chord tones. Naturally, there are strategies for arranging the remaining chord tones that are better than randomness, but they are unnecessary for this demonstration. Repeat this procedure for all pairs of consecutive chord tone objects. For example, our graph traversal for Figure 1 gives the unordered chord tone sets
\begin{equation}
\begin{split}
    [\{(E, 5), (A, 3), (D, 2), (D, 5)\}, \\
    \{(A, 5), (D, 4), (D, 7), (G, 5)\}]
    \end{split}
\end{equation}

We initialize our ordered list to store the arranged chord tones as
\begin{equation}
    [(), (), (), (), (), (), (), ()]
\end{equation}

Our transition tones are $(D, 5)$ for the Amin7 chord and $(D, 4)$ for the D7 chord. Therefore, we set our consecutive transition tones to

\begin{equation}
    [(), (), (), (D, 5), (D, 4), (), (), ()]
\end{equation}

Then, we randomly arrange the other notes in their chord tone slots, for example

\begin{equation}
\begin{split}
    [(D, 2), (E, 5), (A, 3), (D, 5), \\ (D, 4), (A, 5), (D, 7), (G, 5)]
    \end{split}
\end{equation}

Figure 2 renders this chord tone line as tablature, which would be the desired output for this system.

\begin{figure}[h]
\centering
\includegraphics[width=0.9\columnwidth]{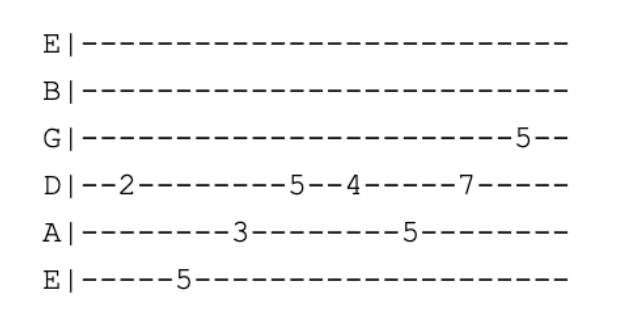}
\caption{The chord tone line from equation (6) as guitar tablature.}
\end{figure}

\subsection{Introducing slight randomness to the output}

This graph engine has a deterministic output. The output tablature will be constant if a user queries the engine with the chord progression Amin7 to D7. Consistent behavior is suboptimal for our educational use case because a guitar student should practice chord tones in many positions for a given chord progression. We can nudge the model towards different neck positions while keeping the weight constraints satisfied by giving a random value to the source node, rather than NULL.

Currently, our source block has a NULL value and our weight constraint is 0 for any edge connected to the source block. We can introduce randomness by setting our source block to a random arpeggio. One such arpeggio is a single note on a random position of the guitar fretboard. If we call that one note $n\in C_0$, Equation 1 simplifies to

\begin{equation}
w(C_0, C_1) = \min_{i \in C_1} |n - i|
\end{equation}

Equation 6 gives the weight only between the Source node and the first layer of nodes. The weight will have the lowest value when any chord tones for the first chord in the progression are closest to the random starting position.

Now, when we query the chord graph, the optimal path is slightly randomized in favor of the randomized start position. 

\section{Use as an Educational Tool}\label{sec:use_in_eduction}

Our chord tone engine takes chord names, such as Amin7, and generates feasible chord tone suggestions for practice. The primary goal of this generation engine is to increase access to jazz guitar education by allowing a guitar student to use the engine for suggestions in their practice. Therefore, this engine requires a software system design to make the educational tool robust and accessible. Such a system needs an input interface, where the user can query the system with any chord progression. After the input is fed through the generation engine, the resulting chord tones must be displayed in a method that is clear to the user and complies with standard guitar notation. Figure 3 shows a skeleton for this system design.

\begin{figure}[h]
\centering
\includegraphics[width=0.9\columnwidth]{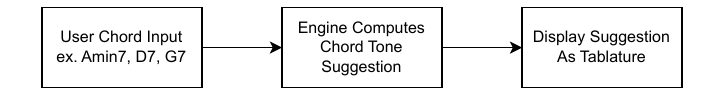}
\caption{A skeleton system for guitar chord input, efficient guitar tablature computation, and guitar chord-tone soloing output.\label{fig:example}}
\end{figure}

The system could accept chord input through various user interfaces. In the most basic system design, the user could input chord names directly as text. The system could also utilize automatic chord recognition from audio (ACR), similar to the methods described in Pauwels et al. \cite{ACR:19}. Such a system could parse an audio file of chords in a song and output chord tone soloing examples for the guitar student. Alternatively, the system could utilize optical music recognition techniques (OMR), such as automatic chord symbol recognition used by Kodirov et al. \cite{Kodirov:14}, for physical chord charts, such as the Real Book of Jazz. A guitar student could include our system in their standard practice routine by taking a picture of any chord chart and generating example chord tones with which to practice. An ideal system would have all three options, input from text, audio, or picture, for the best user experience.

Our system outputs tablature notation, which specifies an exact string and fret position to play a note. Guitarists typically read two types of notation: staff and tablature. Staff notation is the standard musical notation for many Western instruments, with five lines and four spaces. Staff notation can be used for chord tones but does not describe an exact position. For instance, the C note on the second highest space on the treble clef can be played in at least five positions on the guitar. Tablature notation has six lines for the standard six strings on the guitar, and each note is written with a number representing the fret position on the string line. So, if a guitarist reads tablature with a 10 written on the second lowest line, they can only play that note as the G note on the 10th fret of the A string. Therefore, tablature is the clear notation technique for demonstrating the exact position where the guitar student should play each chord tone. Tablature is straightforward to render on a website or application, needing six lines drawn per row of tablature, and the chord tones rendered as fret positions.

%\begin{equation}
%E=mc^{2}.
%\label{eq:Emc2}
%\end{equation}

%\begin{table}[h]
% \begin{center}
% \begin{tabular}{|l|l|}
%  \hline
%  String Value & Numeric value \\
%  \hline
%  Hello ICMC2025 & 2025 \\
%  \hline
% \end{tabular}
%\end{center}
% \caption{Table captions should be placed below the table.}
% \label{tab:example}
%\end{table}

%\begin{figure}[h]
%\centering
%\includegraphics[width=0.9\columnwidth]{figure.eps}
%\caption{Figure captions are placed below the figure, exactly like this. Please note that the printed proceedings will be black and white.\label{fig:example}}
%\end{figure}

%\footnote{This is a footnote.}

%\newpage

\section{Conclusions}
This work presents an engine for generating chord tone soloing suggestions for guitar students. The engine accepts chord names as input and outputs tablature for the chord tone suggestions. The engine constructs a weighted graph where nodes represent a set of chord tones for a chord in the progression and edges represent the movement between consecutive chords in a progression. The edge weight is calculated by quantifying desirable metrics when transitioning between sets of chord tones, most notably the quality of a transition tone. Then, a shortest-path search of the graph and a reconstruction yields a playable chord tone line. We present a larger system design for a full educational tool that promotes ease of use and unambiguous output.

Future work could improve the graph engine's weights. Currently, our graph engine calculates weights using a deterministic method that applies a strict set of rules. Machine learning approaches could be explored to calculate weights based on the tendencies of professional jazz guitarists. This improvement would involve curating a dataset of jazz guitarist chord tone solos to train a model. Further future work could be done on creating a full end-to-end learning system for students to use in their practice. For instance, the system could leverage the ACR and OMR techniques discussed in section 3 or study how students interact with such a system in a human-computer interaction paradigm. Chord tone soloing is a core part of learning jazz guitar improvisation and is difficult to learn given the high-cost barrier with professional guitar lessons and the lack of practice materials online. This work will lower that barrier and allow students to access generated examples for any chord progression.

%%%%%%%%%%%%%%%%%%%%%%%%%%%%%%%%%%%%%%%%%%%%%%%%%%%%%%%%%%%%%%%%%%%%%%%%%%%%%
%bibliography here
\bibliography{icmc2025_paper_template}

\end{document}